\pdfoutput=1

\documentclass[11pt]{article}

\usepackage[]{acl}

\usepackage{times}
\usepackage{latexsym}

\usepackage[T1]{fontenc}

\usepackage[utf8]{inputenc}

\usepackage{microtype}

\usepackage{multirow}
\usepackage{graphicx}
\usepackage{inconsolata}
\usepackage{enumitem}
\usepackage{amsmath,amssymb} 
\usepackage{fdsymbol}
\usepackage{booktabs}
\usepackage{amsfonts}
\usepackage{tabularx}               
\usepackage{xcolor}

\usepackage{xparse}
\NewDocumentCommand{\heng}
{ mO{} }{\textcolor{red}{\textsuperscript{\textit{Heng}}\textsf{\textbf{\small[#1]}}}}

\NewDocumentCommand{\cheng}
{ mO{} }{\textcolor{orange}{\textsuperscript{\textit{Cheng}}\textsf{\textbf{\small[#1]}}}}

\title{Ensemble Transfer Learning for Multilingual Coreference Resolution}

\author{Tuan Lai, Heng Ji\\
	    Department of Computer Science\\
	    University of Illinois Urbana-Champaign\\
        \{tuanml2, hengji\}@illinois.edu
}

\begin{document}
\maketitle

\begin{abstract}
Entity coreference resolution is an important research problem with many applications, including information extraction and question answering. Coreference resolution for English has been studied extensively. However, there is relatively little work for other languages. A problem that frequently occurs when working with a non-English language is the scarcity of annotated training data. To overcome this challenge, we design a simple but effective ensemble-based framework that combines various transfer learning (TL) techniques. We first train several models using different TL methods. Then, during inference, we compute the unweighted average scores of the models' predictions to extract the final set of predicted clusters. Furthermore, we also propose a low-cost TL method that bootstraps coreference resolution models by utilizing Wikipedia anchor texts. Leveraging the idea that the coreferential links naturally exist between anchor texts pointing to the same article, our method builds a sizeable distantly-supervised dataset for the target language that consists of tens of thousands of documents. We can pre-train a model on the pseudo-labeled dataset before finetuning it on the final target dataset. Experimental results on two benchmark datasets, OntoNotes and SemEval, confirm the effectiveness of our methods. Our best ensembles consistently outperform the baseline approach of simple training by up to 7.68\% in the F1 score. These ensembles also achieve new state-of-the-art results for three languages: Arabic, Dutch, and Spanish\footnote{Data and code will be made available upon publication.}.
\end{abstract} 


\section{Introduction}

Within-document entity coreference resolution is the process of clustering entity mentions in a document that refer to the same entities \cite{jietal2005using,luozitouni2005multi,ng2010supervised,Ng17a}. It is an important research problem, with applications in various downstream tasks such as entity linking \cite{lingetal2015design,kunduetal2018neural}, question answering \cite{dhingraetal2018neural}, and dialog systems \cite{gaoetal2019interconnected}. Researchers have recently proposed many neural methods for coreference resolution, ranging from span-based end-to-end models \cite{leeetal2017end,leeetal2018higher} to formulating the task as a question answering problem \cite{wuetal2020corefqa}. Given enough annotated training data, deep neural networks can learn to extract useful features automatically. As a result, on English benchmarks with abundant labeled training documents, the mentioned neural methods consistently outperform previous handcrafted feature-based techniques \cite{raghunathanetal2010multi,10.1162/COLI_a_00152}, achieving new state-of-the-art (SOTA) results.

\begin{figure*}[!ht]
\centering
\includegraphics[width=\textwidth]{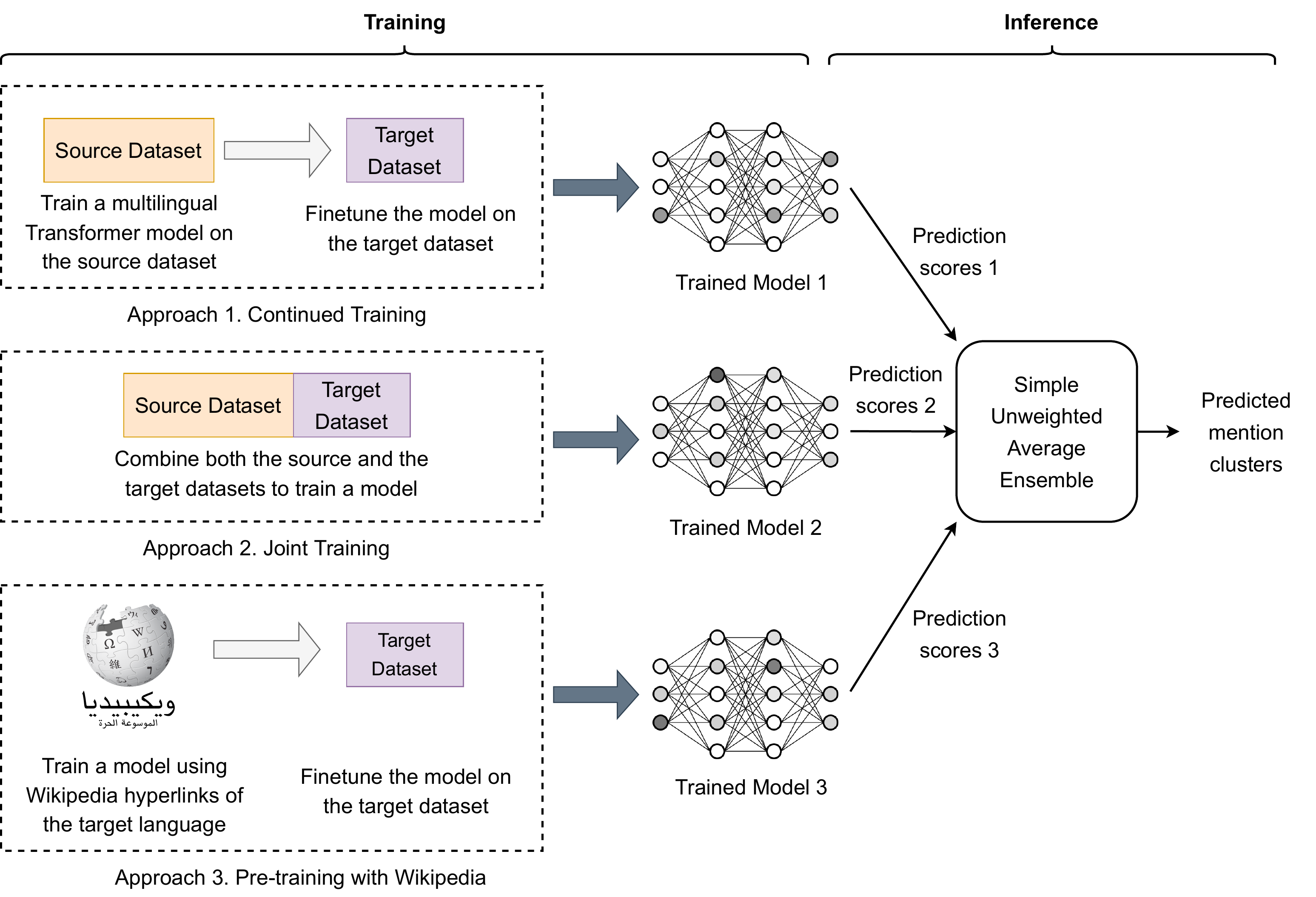}
\caption{An overview of our framework. We first train several coreference resolution models using different TL approaches. During inference, we use a simple unweighted averaging method to combine the models' predictions.}
\label{fig:task}
\end{figure*}

Compared to the amount of research on English coreference resolution, there is relatively little work for other languages. A problem that frequently occurs when working with a non-English language is the scarcity of annotated training data. For example, the benchmark OntoNotes dataset contains about eight times more documents in English than in Arabic \cite{pradhanetal2012conll}. Some recent studies aim to overcome this challenge by applying standard cross-lingual transfer learning (TL) methods such as continued training or joint training \cite{kunduetal2018neural,prazaketal2021multilingual}. In continued training, a model pretrained on a source dataset is further finetuned on a (typically smaller) target dataset \cite{xiavandurme2021moving}. In joint training, a model is trained on the concatenation of the source and target datasets \cite{min2021exploring}. The mentioned studies only use one transfer method at a time, and they do not explore how to combine multiple TL techniques effectively. This can be sub-optimal since different learning methods can be complementary \cite{Liu2019MultiTaskDN,lietal2021taskadaptive}. For example, our experimental results to be discussed later show that continued training and joint training are highly complementary. Furthermore, a disadvantage of using a cross-lingual transfer method is the requirement of a labeled coreference resolution dataset in some source language (usually English).


In this work, we propose an effective ensemble-based framework for combining various TL techniques. We first train several coreference models using different TL methods. During inference, we compute the unweighted average scores of the models' predictions to extract the final set of mention clusters. We also propose a low-cost TL method that bootstraps coreference models without using a labeled dataset in some source language. The basic idea is that the coreference relation often holds between anchor texts pointing to the same Wikipedia article. Based on this observation, our TL method builds a sizable distantly-supervised dataset for the target language from Wikipedia. We can then pre-train a model on the pseudo-labeled dataset before finetuning it on the final target dataset. Experimental results on two datasets, OntoNotes and SemEval \cite{recasensetal2010semeval}, confirm the effectiveness of our proposed methods. Our best ensembles outperform the baseline approach of simple training by up to 7.68\% absolute gain in the F1 score. These ensembles also achieve new SOTA results for three languages: Arabic, Dutch, and Spanish.

In summary, our main contributions include:
\begin{itemize}[nosep]
    \item We introduce an ensemble-based framework that combines various TL methods effectively.
    \item We design a new TL method that leverages Wikipedia to bootstrap coreference models.
    \item Extensive experimental results show that our proposed methods are highly effective and provide useful insights into entity coreference resolution for non-English languages.
\end{itemize}




\section{Methods}
Figure \ref{fig:task} shows an overview of our framework. During the training stage, we train several coreference resolution models using various TL approaches. For simplicity, we use the same span-based architecture (Section \ref{sec:baseline_archi}) for every model to be trained. However, starting from the same architecture, using different learning methods typically results in models with different parameters. In this work, our framework uses two types of TL methods: (a) cross-lingual TL approaches (Section \ref{sec:cross_lingual_tl}) and (b) our newly proposed Wikipedia-based approach (Section \ref{sec:wikipedia_pretraining}). The cross-lingual TL methods require a labeled coreference resolution dataset in some source language, but our Wikipedia-based method does not have that limitation. Our framework is general as it can work with other learning methods (e.g., self-distillation). During inference, we use a simple unweighted averaging method to combine the trained models' predictions (Section \ref{sec:ensemble_approach}).
\subsection{Span-based End-to-End Coreference Resolution}\label{sec:baseline_archi}
In this work, the architecture of every model is based on the popular span-based \textit{e2e-coref} model \cite{leeetal2017end}. Given an input document consisting of $n$ tokens, our model first forms a contextualized representation for each input token using a multilingual Transformer encoder such as XLM-R \cite{Conneau2020UnsupervisedCR}. Let $\textbf{X} = (\textbf{x}_1, ..., \textbf{x}_n)$ be the output of the encoder. For each candidate span $i$, we define its representation $\textbf{g}_i$ as:
\begin{equation}
    \textbf{g}_i = [\textbf{x}_{\text{START(i)}}, \textbf{x}_{\text{END(i)}}, \hat{\textbf{x}}_{i}, \phi(s_i)]
\end{equation}
where $\text{START}(i)$ and $\text{END}(i)$ denote the start and end indices of span $i$ respectively. $\hat{\textbf{x}}_{i}$ is an attention-weighted sum of the token representations in the span \cite{leeetal2017end}. $\phi(s_i)$ is a feature vector encoding the size of the span.

To maintain tractability, we only consider spans with up to $L$ tokens. The value of $L$ is selected empirically and set to be 30. All the span representations are fed into a mention scorer $s_m(.)$:
\begin{equation} \label{equ:mention_score}
    s_m(i) = \text{FFNN}_{m}(\textbf{g}_i)
\end{equation}
where $\text{FFNN}_{m}$ is a feedforward neural network with ReLU activations. Intuitively, $s_m(i)$ indicates whether span $i$ is indeed an entity mention. 

After scoring the spans using $\text{FFNN}_{m}$, we only keep spans with high mention scores\footnote{We describe the exact filtering criteria in Section \ref{sec:data_and_experiment_setup}.}. We denote the set of the unpruned spans as $S$. Then, for each remaining span $i \in S$, the model predicts a distribution $\hat{P}(j)$ over its antecedents\footnote{All spans are ordered based on their start indices. Spans with the same start index are ordered by their end indices.} $j \in Y(i)$:
\begin{equation} \label{equ:pred_dist}
\begin{split}
\hat{P}(j) &= \frac{\exp{(s(i,j))}}{\sum_{k \in Y(i)} \exp{(s(i,k))}} \\[0.8em]
s(i, j) &= \text{FFNN}_{s}\big(\big\lbrack \textbf{g}_i, \textbf{g}_j, \textbf{g}_i \circ \textbf{g}_j, \phi(i,j) \big\rbrack \big)
\end{split}
\end{equation}
where $Y(i) = \{\epsilon, 1, ..., i-1\}$ is a set consisting of a dummy antecedent $\epsilon$ and all spans that precede $i$. The dummy antecedent $\epsilon$ represents two possible cases: (1) the span $i$ is not an entity mention, or (2) the span $i$ is an entity mention, but it is not coreferential with any remaining preceding span. $\text{FFNN}_{s}$ is a feedforward network, and $\circ$ is element-wise multiplication. $\phi(i,j)$ encodes the distance between the two spans $i$ and $j$. Finally, note that $s(i, \epsilon)$ is fixed to be 0.

Given a labeled document $D$ and a model with parameters $\theta$, we define the mention detection loss:
\begin{align*}
\mathcal{L}_{detect}(\theta, D) &= -\frac{1}{|S|}\sum_{i \in S} \mathcal{L}_{detect}(\theta, i) \\
\mathcal{L}_{detect}(\theta, i) &= y_i \log{\hat{y_i}} + (1 - y_i) \log{(1 - \hat{y_i})}
\end{align*}
where $\hat{y_i} = \text{sigmoid}(s_m(i))$, and $y_i = 1$ if and only if span $i$ is in one of the gold-standard  mention clusters. In addition, we also want to maximize the marginal log-likelihood of all correct antecedents implied by the gold-standard  clustering:
\begin{align*}
    \mathcal{L}_{cluster}(\theta, D) = -\log{\prod_{i \in S} \sum_{\hat{y} \in Y(i) \cap \text{GOLD}(i)} \hat{P}(\hat{y})}
\end{align*}
where $\text{GOLD}(i)$ are gold antecedents for span $i$. $\hat{P}(\hat{y})$ is calculated using Equation \ref{equ:pred_dist}. Our final loss combines mention detection and clustering:
\begin{equation} \label{loss_eq}
\mathcal{L}(\theta, D) = \mathcal{L}_{detect}(\theta, D) + \mathcal{L}_{cluster}(\theta, D)
\end{equation}
\subsection{Cross-Lingual Transfer Learning} \label{sec:cross_lingual_tl}

Inspired by previous studies \cite{xiavandurme2021moving,min2021exploring,prazaketal2021multilingual}, we investigate two different cross-lingual transfer learning methods: \textit{continued training} and \textit{joint training}. Both methods assume the existence of a labeled dataset in some source language. In this work, we use the English OntoNotes dataset \cite{pradhanetal2012conll} as the source dataset, as it contains nearly 3,500 annotated documents (Table \ref{tab:dataset-stats}).

\begin{figure*}[!ht]
\centering
\includegraphics[width=0.90\textwidth]{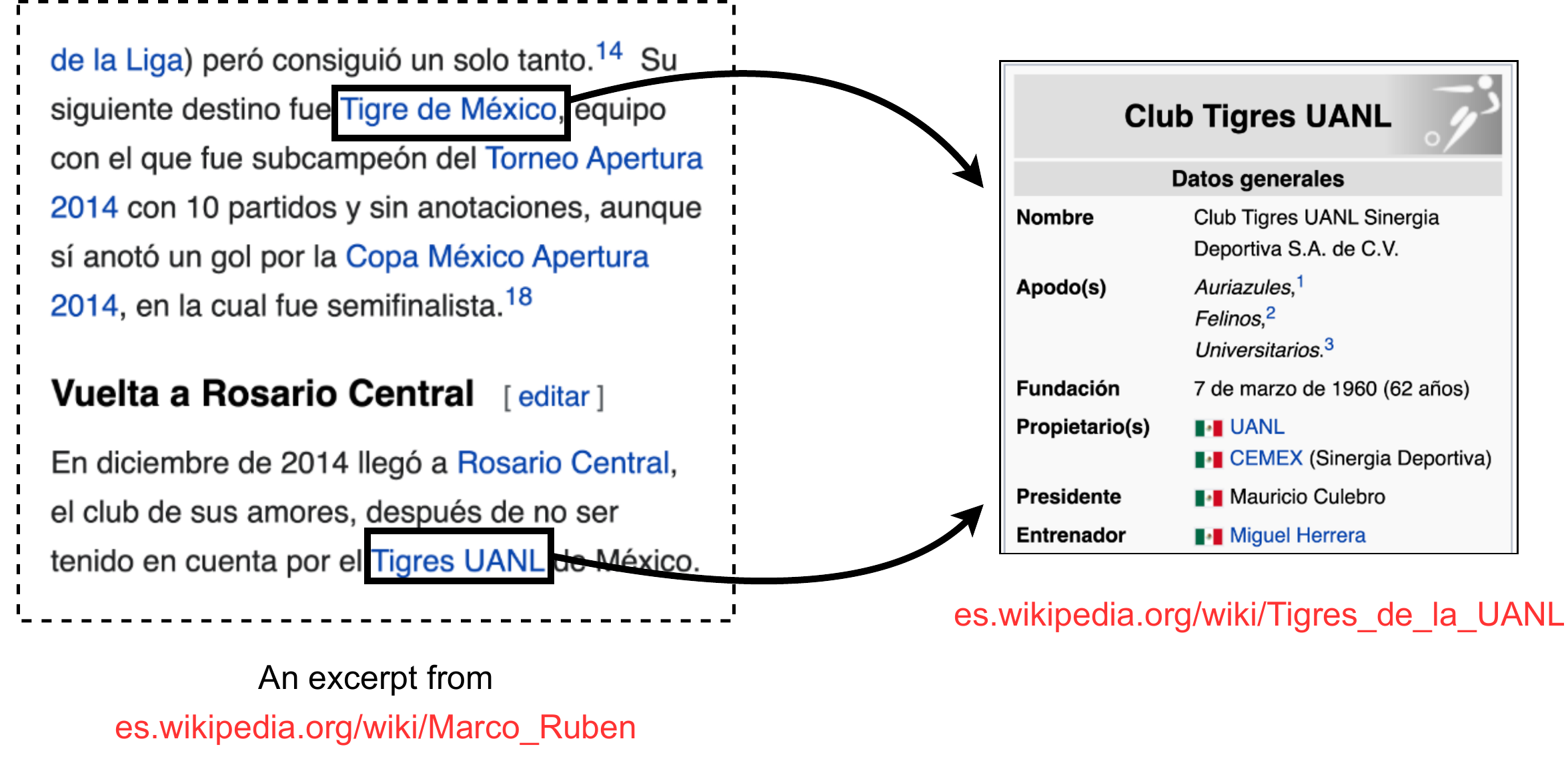}
\caption{Since the hyperlinks of \textit{Tigre de México} and \textit{Tigres UANL} point to the same Wikipedia page, a person who does not know Spanish can still guess that the two mentions are likely to be coreferential. In fact, the two mentions both refer to Tigres UANL, a Mexican professional football club.}
\label{fig:wiki_example}
\end{figure*}

\paragraph{Continued Training.} We first train a coreference resolution model on the source dataset until convergence. After that, we further finetune the pre-trained model on a target dataset. More formally, let $M(f, \theta_0)$ denote an optimization procedure for $f$ with initial guess $\theta_0$. This optimization procedure can, for example, be the application of some stochastic gradient descent algorithm. Also, let $\mathbb{S}$ be the set of all training documents in the source dataset, and let $\mathbb{T}$ denote the set of all training documents in the target dataset. Then, the first stage of continued training can be described as:
\begin{equation}\label{equ:continued_1}
    \hat{\theta}_1 = M\Bigg(\sum_{D \in \mathbb{S}}\mathcal{L}(\theta, D), \hat{\theta}_0\Bigg)
\end{equation}
where $\hat{\theta}_0$ is randomly initialized. Then, the second stage can be described using Equation \ref{equ:continued_2}:
\begin{equation}\label{equ:continued_2}
    \hat{\theta}_2 = M\Bigg(\sum_{D \in \mathbb{T}}\mathcal{L}(\theta, D), \hat{\theta}_1\Bigg)
\end{equation}
Here, $\hat{\theta}_2$ is the parameter set of the final model.

\paragraph{Joint Training.}  We combine both the source and target datasets to train a model. More specifically, using the same notations as above, we can describe joint training by the following equation:
\begin{equation}
    \hat{\theta}_1 = M\Bigg(\sum_{D \in \mathbb{T} \cup \mathbb{S}}\mathcal{L}(\theta, D), \hat{\theta}_0\Bigg)    
\end{equation}
where $\hat{\theta}_0$ is randomly initialized, and $\hat{\theta}_1$ is the parameter set of the final model.
\subsection{Bootstrapping using Wikipedia Hyperlinks} \label{sec:wikipedia_pretraining}

While cross-lingual methods such as continued training and joint training are conceptually simple and typically effective \cite{huang2020cross}, they require the existence of a labeled dataset in some source language. To overcome this limitation, we propose an inexpensive TL method that bootstraps coreference models by utilizing Wikipedia anchor texts. The basic idea is that two anchor texts pointing to the same Wikipedia page are likely coreferential (See Figure \ref{fig:wiki_example} for an example). Our method builds a large distantly-supervised dataset $\mathbb{W}$ for the target language by leveraging this observation:
\begin{equation}
\mathbb{W} = \{D_1, D_2, ..., D_m\}
\end{equation}
where $D_i$ is a text document constructed from some Wikipedia page written in the target language. The number of mentions in $D_i$ is the same as the number of anchor texts in the text portion of the original Wikipedia article. We consider two mentions in $D_i$ to be coreferential if and only if their corresponding anchor texts point to the same article.

After constructing $\mathbb{W}$, we follow a two-step process similar to the continued training approach. We first train a conference resolution model on $\mathbb{W}$ until convergence. Then, we finetune the pre-trained model on the final target dataset.


Compared to a manually-labeled dataset, $\mathbb{W}$ has several disadvantages. Not all entity mentions are exhaustively marked in Wikipedia documents. For example, in Spanish Wikipedia, pronouns are typically not annotated. Nevertheless, since Wikipedia is one of the largest multilingual repositories for information, $\mathbb{W}$ is generally large (see Table \ref{tab:dataset-stats} for some statistics), and it contains documents on various topics. As such, $\mathbb{W}$ can still provide some useful distant supervision signals, and so it can serve as a source dataset in the TL process.
\subsection{Ensemble-Based Coreference Resolution} \label{sec:ensemble_approach}
During the training stage, we train three different coreference resolution models using the TL approaches described above. At test time, we use a simple unweighted averaging method to combine the models’ predictions. More specifically, for a candidate span $i$ with no more than $L$ tokens, we compute its mention score as follows:
\begin{equation}\label{equ:ensembled_mention_score}
    s_{m,\text{ensemble}}(i) = \frac{\big(s_{m,1}(i) + s_{m,2}(i) + s_{m,3}(i)\big)}{3}
\end{equation}
where $s_{m,1}(i)$, $s_{m,2}(i)$, and $s_{m,3}(i)$ are the mention scores produced by the three trained models separately (refer to Equation \ref{equ:mention_score}). Intuitively, these scores indicate whether span $i$ is an entity mention. 


Similar to the process described in Section \ref{sec:baseline_archi}, after scoring every span whose length is no more than $L$ using Equation \ref{equ:ensembled_mention_score}, we only keep spans with high mention scores\footnote{We describe the exact filtering criteria in Section \ref{sec:data_and_experiment_setup}.}. Then, for each remaining span $i$, we predict a distribution over its antecedents $j \in Y(i)$ as follows:
\begin{equation} \label{sec:ensembled_pred_dist}
\begin{split}
\hat{P}_{\text{ensemble}}(j) &= \frac{\exp{\big(s_{\text{ensemble}}(i,j)}\big)}{\sum_{k \in Y(i)} \exp{\big(s_{\text{ensemble}}(i,k)}\big)} \\[0.8em]
s_{\text{ensemble}}(i, j) &= \frac{s_1(i,j) + s_2(i,j) + s_3(i, j)}{3}
\end{split}
\end{equation}
where $s_1(i,j)$, $s_2(i,j)$, and $s_3(i,j)$ are the pairwise scores produced by the trained models separately (Eq. \ref{equ:pred_dist}). We fix $s_{\text{ensemble}}(i, \epsilon)$ to be 0.

After computing the antecedent distribution for each remaining span, we can extract the final set of mention clusters. Note that while we consider only three individual TL methods in this work, Equation \ref{equ:ensembled_mention_score} and Equation \ref{sec:ensembled_pred_dist} can easily be extended for the case when we use more TL methods.

\section{Experiments}
\subsection{Data and Experiments Setup}\label{sec:data_and_experiment_setup}
\paragraph{Evaluation metrics} Following prior work \cite{pradhanetal2012conll}, we evaluate coreference using the average $F_1$ between $\text{B}^3$ \cite{Bagga1998AlgorithmsFS}, MUC \cite{Vilain1995AMC}, and $\text{CEAF}_{\phi_4}$ \cite{Luo2005OnCR}. We refer to this metric as AVG.

\renewcommand{\arraystretch}{1.2}
\begin{table}[!ht]
\centering
\resizebox{\linewidth}{!}{%
\begin{tabular}{l|rrr}
\hline
Dataset & Training & Dev & Test \\ \hline
\textit{Source Datasets} & & & \\
OntoNotes English & 2,802 & 343 & 348 \\
Wikipedia-based Arabic Dataset & 64,850 & 250 & 250 \\
Wikipedia-based Dutch Dataset & 46,715 & 250 & 250 \\
Wikipedia-based Spanish Dataset & 104,520 & 250 & 250 \\\hline
\textit{Target Datasets} & & & \\
OntoNotes Arabic & 359 & 44 & 44 \\
SemEval Dutch & 145 & 23 & 72 \\
SemEval Spanish & 875 & 140 & 168 \\ \hline
\end{tabular}%
}
\caption{Number of documents for each of the datasets.}
\label{tab:dataset-stats}
\vspace{-6mm}
\end{table}

\paragraph{Datasets} Table \ref{tab:dataset-stats} shows the basic statistics of all the datasets we used in this work. When using a cross-lingual TL method, we use the English portion of OntoNotes \cite{pradhanetal2012conll} as the source dataset. We explore three target datasets: OntoNotes Arabic \cite{pradhanetal2012conll}, SemEval Dutch \cite{recasensetal2010semeval}, and SemEval Spanish\cite{recasensetal2010semeval}. These datasets contain data in three different languages.

OntoNotes does not annotate singleton mentions (i.e., noun phrases not involved in any coreference chain). It only has annotations for non-singleton mentions. SemEval has annotations for singletons.

\paragraph{Wikipedia-based Dataset Construction} To construct a distantly-supervised dataset, we first download a complete Wikipedia dump in the target language. We then extract clean text and hyperlinks from the dump using WikiExtractor\footnote{\url{https://tinyurl.com/wikiextractor}}. For each preprocessed article, we cluster its anchor texts based on the destinations of their hyperlinks. We also filter out articles with too few coreference links (e.g., articles that only have singleton mentions). 

\renewcommand{\arraystretch}{1.25}
\begin{table*}[!ht]
\centering
\resizebox{\textwidth}{!}{%
\small
\begin{tabular}{lccc}
\hline
 & Arabic & Dutch & Spanish \\ \hline
\textit{Baselines} & & &  \\
$\mdblkdiamond$ Previous SOTA (Table \ref{tab:previous-sota}) & 64.55 & 55.40 & 51.30 \\
$\mdblkdiamond$ Baseline Approach (trained using the target dataset) & 63.70 & 52.81 & 72.18 \\ \hline
\textit{Individual Transfer/Pretraining Methods (Sections \ref{sec:cross_lingual_tl} and \ref{sec:wikipedia_pretraining})} & & &  \\
$\mdblksquare$ Continued Training & 64.96 & 58.90  & 74.05 \\
$\mdblksquare$ Joint Training & 65.50 & 58.76 & 73.53 \\
$\mdblksquare$ Wikipedia Pre-Training & 63.78 & 53.15 & 73.35 \\ \hline
\textit{Ensembles (Section \ref{sec:ensemble_approach})} & & &  \\
$\mdblklozenge$ Three models, each trained using the baseline approach & 64.70 & 54.44 &  73.35 \\
$\mdblklozenge$ Baseline Approach $\oplus$ Wikipedia Pre-Training & 65.75 & 55.25 & 74.19 \\
$\mdblklozenge$ Joint Training $\oplus$ Wikipedia Pre-Training & 66.63 & 58.18 & 74.82 \\
$\mdblklozenge$ Continued Training $\oplus$ Wikipedia Pre-Training & 66.24 & 57.88 & 75.43 \\
$\mdblklozenge$ Continued Training $\oplus$ Joint Training & 65.79 & \textbf{60.49} & 74.93 \\
$\mdblklozenge$ Continued Training $\oplus$ Joint Training $\oplus$ Wikipedia Pre-Training & \textbf{66.72} & 59.66 & \textbf{75.62} \\
\hline
\textit{Oracle-Guided Ensembles (Section \ref{sec:analysis_oracle})} & & & \\
$\mdlgwhtdiamond$ Continued Training $\oplus$ Joint Training $\oplus$ Wikipedia Pre-Training & 77.53 & 75.19 & 83.12 \\\hline
\end{tabular}%
}
\caption{Overall $F_1$  (in \%) on OntoNotes Arabic, SemEval Dutch, and SemEval Spanish.}
\label{tab:overall_results}
\end{table*}

\renewcommand{\arraystretch}{1.2}
\begin{table*}[ht!]
\centering
\resizebox{\textwidth}{!}{%
\begin{tabular}{lllcc}
\hline
Dataset & Prior Work & Approach & Prev. Score & Our Best \\ \hline
OntoNotes Arabic  & \cite{min2021exploring} & GigaBERT + C2F + Joint Training & 64.55 & \textbf{66.72} \\
SemEval Dutch & \cite{xiavandurme2021moving} & XLM-R + ICoref + Continued Training & 55.40 & \textbf{60.49}  \\
SemEval Spanish & \cite{xiavandurme2021moving} & XLM-R + ICoref + Continued Training & 51.30 & \textbf{75.62}  
\end{tabular}%
}
\caption{Test $F_1$ (in \%) on the target datasets and the previous SOTA on each dataset (to the best of our knowledge).}
\label{tab:previous-sota}
\end{table*}

\paragraph{General Hyperparameters} We 
use two different learning rates, one for the lower pretrained Transformer encoder and one for the upper layers. For every setting, the lower learning rate is 1e-5, the upper learning rate is 1e-4, and the span length limit $L$ is 30. The number of training/pre-training epochs is set to be 25 in most cases. When pre-training a model on a Wikipedia-based dataset, the number of epochs is 5. When fine-tuning a model already pre-trained on Dutch Wikipedia or Spanish Wikipedia, the number of epochs is 50. During each training/pre-training process, we pick the checkpoint which achieves the best AVG score on the appropriate dev set as the final checkpoint.

\paragraph{Transformer Encoders} When the target dataset is OntoNotes Arabic, we use GigaBERT \cite{lanetal2020empirical} as the Transformer encoder. GigaBERT is an English-Arabic bilingual language model pre-trained from the English and Arabic Gigaword corpora. When the target dataset is SemEval Dutch or SemEval Spanish, we use the multilingual XLM-RoBERTa (XLM-R) Transformer model \cite{Conneau2020UnsupervisedCR}. More specifically, we use the base version of XLM-R (i.e., \texttt{xlm-roberta-base}).

\paragraph{Span Pruning} As described in Section \ref{sec:baseline_archi}, after computing a mention score for each span whose length is not more than $L$, we only keep spans with high scores. More specifically, when working with a dataset from OntoNotes (e.g., OntoNotes Arabic), we only keep up to $\lambda n$ spans with the highest mention scores \cite{leeetal2017end}. The value of $\lambda$ is selected empirically and set to be 0.18. When working with any other dataset, we keep every span that has a positive mention score.

\subsection{Overall Results} \label{sec:overall_results}
Table \ref{tab:overall_results} shows the overall performance of different approaches. Our baseline approach is to simply train a model with the architecture described in Section \ref{sec:baseline_archi} using only the target dataset of interest. Overall, the performance of a model trained using the baseline approach is positively correlated with the size of the corresponding target dataset, which is expected. A surprising finding is that our baseline approach already outperforms the previous SOTA method for SemEval Spanish \cite{xiavandurme2021moving} by 20.98\% in the F1 score. We speculate that the previous SOTA model for SemEval Spanish is severely undertrained.

Table \ref{tab:overall_results} also shows the results of using different TL methods individually. Each of the TL methods can help improve the coreference resolution performance. While continued training seems to be the most effective approach, it requires the existence of a source dataset (OntoNotes English in this case). On the other hand, our newly proposed Wikipedia-based method can help improve the performance without relying on any labeled source dataset.

Finally, Table \ref{tab:overall_results} also shows the results of using different combinations of learning approaches. Our simple unweighted averaging method is effective across almost all model combinations. In particular, by combining all of the three TL methods discussed previously, we can outperform the previous SOTA methods by large margins. In addition, even without using any labeled source dataset, the combination [Baseline Approach $\oplus$ Wikipedia Pre-Training] can still outperform the previous SOTA methods for Arabic and Spanish. This further confirms the usefulness of our Wikipedia-based TL method. Lastly, combining three models trained using the same baseline approach leads to smaller gains than combining the three TL methods. This is expected as ensemble methods typically work best when the individual learners are diverse \cite{krogh1994neural,Melville2003ConstructingDC}. 

\subsection{Analysis}

\renewcommand{\arraystretch}{1.2}
\begin{table}[!t]
\centering
\resizebox{\linewidth}{!}{%
\begin{tabular}{lc}
\hline
\multicolumn{1}{c}{Approaches} & AVG \\ \hline
Baseline Approach & 40.10 \\
Wikipedia Pre-Training & 42.67 \\
Baseline Approach $\oplus$ Wikipedia Pre-Training & \textbf{45.28} \\\hline
\end{tabular}%
}
\caption{Test F-score (in \%) of various approaches on OntoNotes Arabic when we restrict the size of the gold Arabic training dataset to only 10 documents.}
\label{tab:low_resource_result}
\vspace{-6mm}
\end{table}

\subsubsection{How optimal is our simple unweighted averaging method?} \label{sec:analysis_oracle}

Our averaging approach is equivalent to linear interpolation with equal weights. To analyze the optimality of our method, we compare it to the ``best possible'' interpolation method.

More specifically, we assume that there is an oracle that can tell us which model in an ensemble gives the most accurate prediction for a particular latent variable. Then, for example, suppose we want to score a span $i$ using an ensemble of three models. If $i$ is an entity mention, the oracle will tell us that the model that returns the highest mention score for $i$ is the most accurate. Thus, we can set the score for $i$ to be \(\max{(s_{m,1}(i), s_{m,2}(i), s_{m,3}(i))}\). Following the same logic, if $i$ is not an entity mention, we will set its score to be \(\min{(s_{m,1}(i), s_{m,2}(i), s_{m,3}(i))}\). The same idea can be applied to compute the linking score $s_{\text{ensemble}}(i, j)$ between $i$ and $j$.

In Table \ref{tab:overall_results}, we see a considerable gap between the performance of our simple averaging method and the oracle-guided interpolation method. Therefore, a promising future direction is to experiment with a more context-dependent ensemble method. Nevertheless, our averaging method is simple, and it does not require any further parameter tuning to combine a set of existing models. Finally, the performance of each oracle-guided ensemble is far from perfect, implying that improving the underlying architecture of each model can also be a worthwhile effort.

\subsubsection{How effective is our framework in extremely low-resource settings?}
We conduct experiments on OntoNotes Arabic where we assume that the training dataset for Arabic only has 10 documents and that we do not have any source dataset (Table \ref{tab:low_resource_result}). In this setting, our ensemble substantially outperforms the baseline approach by up to 5.18\% in the F1 score.

\subsubsection{Qualitative Analysis}
We provide some qualitative analyses to demonstrate the strengths of our ensembles in Table \ref{tab:qualitative_examples}.

In the first example, the three highlighted mentions refer to Anna Mas, the director of a center. Our model trained using joint training merged this cluster with a different cluster that refers to a different entity (not shown in the example because of space constraints). In contrast, our models trained using other TL methods did not make that error. As a result, our best ensemble for Spanish predicted the correct cluster for Anna Mas.

The second example is in Dutch. Here, \textit{mannelijk muis} can be translated as \textit{male mouses}, while \textit{hen} can be translated as \textit{them}. Our model trained using continued training failed to extract the mention \textit{mannelijk muis}. Nevertheless, in the end, our ensemble for Dutch was able to extract the mention and correctly link it to the pronoun \textit{hen}.

\renewcommand{\arraystretch}{1.25}
\begin{table}[t!]
\centering
\small
\begin{tabular}{p{0.92\linewidth}}
\hline
... \textcolor{blue}{el director del centro, Anna Mas}, asegurar que el acto pretender ``rechazar el agresión y concienciar a el alumno del incremento de este ataque''. \textcolor{blue}{el director} recordar otro dos agresión ``por llevar hierro dental o el pelo largo''. en mucho ocasión se producir asalto a niño, y el alumno, añadir \textcolor{blue}{Mas}, ``ver como algo normal que les parir por el calle y les quitar el poco dinero que llevar encima'' ...
\\\hline
... Het vakblad Hormones and Behavior beschrijven hoe het voldoend zijn dat \textcolor{blue}{mannelijk muis} een vleug vrouw roken om \textcolor{blue}{hen} weinig bang te maken van kat en wezel ...
\\\hline
\end{tabular}
\caption{\label{tab:qualitative_examples} Examples of mention clusters that were correctly predicted by our ensembles. Blue spans represent coreferential mentions. The first example is in Spanish. The second example is in Dutch.}
\vspace{-6mm}
\end{table}

\section{Related Work}
\subsection{Entity Coreference Resolution}
Recently, neural models for entity coreference resolution have shown superior performance over approaches using hand-crafted features. \newcite{leeetal2017end} proposed the first end-to-end neural coreference resolution model named \textit{e2e-coref}. The model uses a bi-directional LSTM and a head-finding attention mechanism to learn mention representations and calculate mention and antecedent scores. \newcite{leeetal2018higher} extended the \textit{e2e-coref} model by introducing a coarse-to-fine pruning mechanism and a higher-order inference mechanism. The model uses ELMo representations \cite{petersetal2018deep} instead of traditional word embeddings. The model is typically referred to as the \textit{c2f-coref} model. 

Almost all recent studies on entity coreference resolution are influenced by the design of \textit{c2f-coref}. \newcite{joshietal2019bert} built the \textit{c2f-coref} system on top of BERT representations \cite{devlinetal2019bert}. \newcite{feietal2019end} transformed \textit{c2f-coref} into a policy gradient model that can optimize coreference evaluation metrics directly. \newcite{xuchoi2020revealing} studied in depth the higher-order inference (HOI) mechanism of \textit{c2f-coref}. The authors concluded that given a high-performing encoder such as SpanBERT \cite{Joshi2020SpanBERTIP}, the impact of HOI is negative to marginal. Another line of work aims to simplify and/or reduce the computational complexity of \textit{c2f-coref} \cite{xiaetal2020incremental,kirstainetal2021coreference,lai2021end,dobrovolskii2021word}.

The studies mentioned above only trained and evaluated models using English datasets such as OntoNotes English \cite{pradhanetal2012conll} and the GAP dataset \cite{gapdataset}. On the other hand, there is significantly less work on coreference resolution for other languages. For example, while \textit{e2e-coref} was introduced in 2017, the first neural coreference resolver for Arabic was only recently proposed in 2020 \cite{alorainietal2020neural}. For Dutch, many existing systems are still using rule-based \cite{vanCranenburgh2019ADC} or traditional learning-based approaches \cite{hendrickxetal2008coreference,declercqetal2011cross}. Recently, \newcite{pootvancranenburgh2020benchmark} evaluated the performance of \textit{c2f-coref} on Dutch datasets of two different domains: literary novels and news/Wikipedia text.

While our models' architecture is based on \textit{e2e-coref} (Section \ref{sec:baseline_archi}), we go beyond just applying the models to a non-English language in this work. We propose new TL approaches that can take advantage of existing source datasets and Wikipedia to improve the final performance.

\subsection{Transfer Learning for Coreference Resolution}
Compared to English datasets, the size of a coreference resolution dataset for a non-English language is typically smaller. Several recent studies aim to overcome this challenge by applying standard cross-lingual TL methods such as continued training or joint training \cite{kunduetal2018neural,xiavandurme2021moving,prazaketal2021multilingual,min2021exploring}. These studies only use one transfer method at a time, and they do not explore how to combine multiple TL techniques effectively. Our experimental results (Section \ref{sec:overall_results}) show that combining various TL techniques can substantially improve the final coreference resolution performance.

A closely related work by \newcite{yangetal2012domain} proposed an adaptive ensemble method to adapt coreference resolution across domains. Their study did not explicitly focus on improving coreference resolution for non-English languages. In addition, they experimented with the settings where gold standard mentions are assumed to be provided. We do not make that assumption. Each of our models does both mention extraction and linking.
\subsection{Leveraging Wikipedia for Coreference Resolution}
There have been studies on leveraging Wikipedia for coreference resolution. \newcite{eirewetal2021wec} recently created a large-scale cross-document event coreference dataset from  English Wikipedia. For cross-document entity coreference, \newcite{Singh2012WikilinksAL} created Wikilinks by finding hyperlinks to English Wikipedia from a web crawl and using anchor text as mentions. Different from these studies, we focus on within-document entity coreference resolution. In addition, we explore coreference resolution for languages beyond English in this work.

Many previous studies leveraged Wikipedia for related tasks such as name tagging \cite{alotaibilee2012mapping,Nothman2013LearningMN,althobaitietal2014automatic} and entity linking \cite{panetal2017cross,wu2020scalable,decao2020autoregressive}. We leave the extension of our methods to these tasks for future research.

\section{Conclusions and Future Work}
In this work, we propose an ensemble-based framework that combines various TL techniques. We also introduce a low-cost Wikipedia-based TL approach that does not require any labeled source dataset. Our approaches are highly effective, as our best ensembles achieve new SOTA results for three different languages. An interesting future direction is to explore the use of model compression techniques \cite{Hinton2015DistillingTK,Han2016DeepCC,lai2020simple} to reduce the computational complexity of our ensembles.
\section{Limitations}

Multilingual language models such as XLM-R \cite{Conneau2020UnsupervisedCR} and GigaBERT \cite{lanetal2020empirical} are typically pre-trained on large amounts of unlabeled text crawled from the Web. Since these models are optimized to capture the statistical properties of the training data, they tend to pick up on and amplify social stereotypes present in the data \cite{kuritaetal2019measuring}. Since our coreference resolution models use such pre-trained language models, they may also exhibit social biases present on the Web. Identifying and mitigating social biases in neural models is an active area of research \cite{zhaoetal2018gender,sheng2021societal,gupta2022equitable}. In the future, we plan to work on removing social biases from coreference resolution models.

Furthermore, while our proposed methods are highly effective, the performance of our best ensembles is still far from perfect. On OntoNotes Arabic, our best system only achieves an F1 score of 66.72\%. Such performance may not be acceptable for some downstream tasks (e.g., information extraction from critical clinical notes).

Finally, even though Wikipedia is available in more than 300 languages, there are still very few Wikipedia pages for some very rare languages. Our proposed methods are likely to be less effective for such rare languages.

\bibliography{anthology,custom}

\begin{thebibliography}{64}
\expandafter\ifx\csname natexlab\endcsname\relax\def\natexlab#1{#1}\fi

\bibitem[{Aloraini et~al.(2020)Aloraini, Yu, and
  Poesio}]{alorainietal2020neural}
Abdulrahman Aloraini, Juntao Yu, and Massimo Poesio. 2020.
\newblock \href {https://aclanthology.org/2020.crac-1.11} {Neural coreference
  resolution for {A}rabic}.
\newblock In \emph{Proceedings of the Third Workshop on Computational Models of
  Reference, Anaphora and Coreference}, pages 99--110, Barcelona, Spain
  (online). Association for Computational Linguistics.

\bibitem[{Alotaibi and Lee(2012)}]{alotaibilee2012mapping}
Fahd Alotaibi and Mark Lee. 2012.
\newblock \href {https://aclanthology.org/C12-2005} {Mapping {A}rabic
  {W}ikipedia into the named entities taxonomy}.
\newblock In \emph{Proceedings of {COLING} 2012: Posters}, pages 43--52,
  Mumbai, India. The COLING 2012 Organizing Committee.

\bibitem[{Althobaiti et~al.(2014)Althobaiti, Kruschwitz, and
  Poesio}]{althobaitietal2014automatic}
Maha Althobaiti, Udo Kruschwitz, and Massimo Poesio. 2014.
\newblock \href {https://doi.org/10.3115/v1/E14-3012} {Automatic creation of
  {A}rabic named entity annotated corpus using {W}ikipedia}.
\newblock In \emph{Proceedings of the Student Research Workshop at the 14th
  Conference of the {E}uropean Chapter of the Association for Computational
  Linguistics}, pages 106--115, Gothenburg, Sweden. Association for
  Computational Linguistics.

\bibitem[{Bagga and Baldwin(1998)}]{Bagga1998AlgorithmsFS}
Amit Bagga and Breck Baldwin. 1998.
\newblock Algorithms for scoring coreference chains.
\newblock In \emph{The first international conference on language resources and
  evaluation workshop on linguistics coreference}, volume~1, pages 563--566.
  Citeseer.

\bibitem[{Cao et~al.(2021)Cao, Izacard, Riedel, and
  Petroni}]{decao2020autoregressive}
Nicola~De Cao, Gautier Izacard, Sebastian Riedel, and Fabio Petroni. 2021.
\newblock \href {https://openreview.net/forum?id=5k8F6UU39V} {Autoregressive
  entity retrieval}.
\newblock In \emph{International Conference on Learning Representations}.

\bibitem[{Conneau et~al.(2020)Conneau, Khandelwal, Goyal, Chaudhary, Wenzek,
  Guzm{\'a}n, Grave, Ott, Zettlemoyer, and
  Stoyanov}]{Conneau2020UnsupervisedCR}
Alexis Conneau, Kartikay Khandelwal, Naman Goyal, Vishrav Chaudhary, Guillaume
  Wenzek, Francisco Guzm{\'a}n, Edouard Grave, Myle Ott, Luke Zettlemoyer, and
  Veselin Stoyanov. 2020.
\newblock Unsupervised cross-lingual representation learning at scale.
\newblock In \emph{ACL}.

\bibitem[{De~Clercq et~al.(2011)De~Clercq, Hoste, and
  Hendrickx}]{declercqetal2011cross}
Orph{\'e}e De~Clercq, V{\'e}ronique Hoste, and Iris Hendrickx. 2011.
\newblock \href {https://aclanthology.org/R11-1026} {Cross-domain {D}utch
  coreference resolution}.
\newblock In \emph{Proceedings of the International Conference Recent Advances
  in Natural Language Processing 2011}, pages 186--193, Hissar, Bulgaria.
  Association for Computational Linguistics.

\bibitem[{Devlin et~al.(2019)Devlin, Chang, Lee, and
  Toutanova}]{devlinetal2019bert}
Jacob Devlin, Ming-Wei Chang, Kenton Lee, and Kristina Toutanova. 2019.
\newblock \href {https://doi.org/10.18653/v1/N19-1423} {{BERT}: Pre-training of
  deep bidirectional transformers for language understanding}.
\newblock In \emph{Proceedings of the 2019 Conference of the North {A}merican
  Chapter of the Association for Computational Linguistics: Human Language
  Technologies, Volume 1 (Long and Short Papers)}, pages 4171--4186,
  Minneapolis, Minnesota. Association for Computational Linguistics.

\bibitem[{Dhingra et~al.(2018)Dhingra, Jin, Yang, Cohen, and
  Salakhutdinov}]{dhingraetal2018neural}
Bhuwan Dhingra, Qiao Jin, Zhilin Yang, William Cohen, and Ruslan Salakhutdinov.
  2018.
\newblock \href {https://doi.org/10.18653/v1/N18-2007} {Neural models for
  reasoning over multiple mentions using coreference}.
\newblock In \emph{Proceedings of the 2018 Conference of the North {A}merican
  Chapter of the Association for Computational Linguistics: Human Language
  Technologies, Volume 2 (Short Papers)}, pages 42--48, New Orleans, Louisiana.
  Association for Computational Linguistics.

\bibitem[{Dobrovolskii(2021)}]{dobrovolskii2021word}
Vladimir Dobrovolskii. 2021.
\newblock \href {https://doi.org/10.18653/v1/2021.emnlp-main.605} {Word-level
  coreference resolution}.
\newblock In \emph{Proceedings of the 2021 Conference on Empirical Methods in
  Natural Language Processing}, pages 7670--7675, Online and Punta Cana,
  Dominican Republic. Association for Computational Linguistics.

\bibitem[{Eirew et~al.(2021)Eirew, Cattan, and Dagan}]{eirewetal2021wec}
Alon Eirew, Arie Cattan, and Ido Dagan. 2021.
\newblock \href {https://doi.org/10.18653/v1/2021.naacl-main.198} {{WEC}:
  Deriving a large-scale cross-document event coreference dataset from
  {W}ikipedia}.
\newblock In \emph{Proceedings of the 2021 Conference of the North American
  Chapter of the Association for Computational Linguistics: Human Language
  Technologies}, pages 2498--2510, Online. Association for Computational
  Linguistics.

\bibitem[{Fei et~al.(2019)Fei, Li, Li, and Li}]{feietal2019end}
Hongliang Fei, Xu~Li, Dingcheng Li, and Ping Li. 2019.
\newblock \href {https://doi.org/10.18653/v1/P19-1064} {End-to-end deep
  reinforcement learning based coreference resolution}.
\newblock In \emph{Proceedings of the 57th Annual Meeting of the Association
  for Computational Linguistics}, pages 660--665, Florence, Italy. Association
  for Computational Linguistics.

\bibitem[{Gao et~al.(2019)Gao, Li, King, and Lyu}]{gaoetal2019interconnected}
Yifan Gao, Piji Li, Irwin King, and Michael~R. Lyu. 2019.
\newblock \href {https://doi.org/10.18653/v1/P19-1480} {Interconnected question
  generation with coreference alignment and conversation flow modeling}.
\newblock In \emph{Proceedings of the 57th Annual Meeting of the Association
  for Computational Linguistics}, pages 4853--4862, Florence, Italy.
  Association for Computational Linguistics.

\bibitem[{Gupta et~al.(2022)Gupta, Dhamala, Kumar, Verma, Pruksachatkun,
  Krishna, Gupta, Chang, Steeg, and Galstyan}]{gupta2022equitable}
Umang Gupta, Jwala Dhamala, Varun Kumar, Apurv Verma, Yada Pruksachatkun,
  Satyapriya Krishna, Rahul Gupta, Kai-Wei Chang, Greg~Ver Steeg, and Aram
  Galstyan. 2022.
\newblock Mitigating gender bias in distilled language models via
  counterfactual role reversal.
\newblock In \emph{ACL Finding}.

\bibitem[{Han et~al.(2016)Han, Mao, and Dally}]{Han2016DeepCC}
Song Han, Huizi Mao, and William~J. Dally. 2016.
\newblock Deep compression: Compressing deep neural network with pruning,
  trained quantization and huffman coding.
\newblock \emph{arXiv: Computer Vision and Pattern Recognition}.

\bibitem[{Hendrickx et~al.(2008)Hendrickx, Bouma, Coppens, Daelemans, Hoste,
  Kloosterman, Mineur, Van Der~Vloet, and
  Verschelde}]{hendrickxetal2008coreference}
Iris Hendrickx, Gosse Bouma, Frederik Coppens, Walter Daelemans, Veronique
  Hoste, Geert Kloosterman, Anne-Marie Mineur, Joeri Van Der~Vloet, and
  Jean-Luc Verschelde. 2008.
\newblock \href
  {http://www.lrec-conf.org/proceedings/lrec2008/pdf/49_paper.pdf} {A
  coreference corpus and resolution system for {D}utch}.
\newblock In \emph{Proceedings of the Sixth International Conference on
  Language Resources and Evaluation ({LREC}'08)}, Marrakech, Morocco. European
  Language Resources Association (ELRA).

\bibitem[{Hinton et~al.(2015)Hinton, Vinyals, and
  Dean}]{Hinton2015DistillingTK}
Geoffrey~E. Hinton, Oriol Vinyals, and Jeffrey Dean. 2015.
\newblock Distilling the knowledge in a neural network.
\newblock \emph{ArXiv}, abs/1503.02531.

\bibitem[{Huang et~al.(2020)Huang, Kuchaiev, O'Neill, Lavrukhin, Li, Flores,
  Kucsko, and Ginsburg}]{huang2020cross}
Jocelyn Huang, Oleksii Kuchaiev, Patrick O'Neill, Vitaly Lavrukhin, Jason Li,
  Adriana Flores, Georg Kucsko, and Boris Ginsburg. 2020.
\newblock Cross-language transfer learning, continuous learning, and domain
  adaptation for end-to-end automatic speech recognition.
\newblock \emph{arXiv preprint arXiv:2005.04290}.

\bibitem[{Ji et~al.(2005)Ji, Westbrook, and Grishman}]{jietal2005using}
Heng Ji, David Westbrook, and Ralph Grishman. 2005.
\newblock \href {https://aclanthology.org/H05-1003} {Using semantic relations
  to refine coreference decisions}.
\newblock In \emph{Proceedings of Human Language Technology Conference and
  Conference on Empirical Methods in Natural Language Processing}, pages
  17--24, Vancouver, British Columbia, Canada. Association for Computational
  Linguistics.

\bibitem[{Joshi et~al.(2020)Joshi, Chen, Liu, Weld, Zettlemoyer, and
  Levy}]{Joshi2020SpanBERTIP}
Mandar Joshi, Danqi Chen, Yinhan Liu, Daniel~S. Weld, Luke Zettlemoyer, and
  Omer Levy. 2020.
\newblock Spanbert: Improving pre-training by representing and predicting
  spans.
\newblock \emph{Transactions of the Association for Computational Linguistics},
  8:64--77.

\bibitem[{Joshi et~al.(2019)Joshi, Levy, Zettlemoyer, and
  Weld}]{joshietal2019bert}
Mandar Joshi, Omer Levy, Luke Zettlemoyer, and Daniel Weld. 2019.
\newblock \href {https://doi.org/10.18653/v1/D19-1588} {{BERT} for coreference
  resolution: Baselines and analysis}.
\newblock In \emph{Proceedings of the 2019 Conference on Empirical Methods in
  Natural Language Processing and the 9th International Joint Conference on
  Natural Language Processing (EMNLP-IJCNLP)}, pages 5803--5808, Hong Kong,
  China. Association for Computational Linguistics.

\bibitem[{Kirstain et~al.(2021)Kirstain, Ram, and
  Levy}]{kirstainetal2021coreference}
Yuval Kirstain, Ori Ram, and Omer Levy. 2021.
\newblock \href {https://doi.org/10.18653/v1/2021.acl-short.3} {Coreference
  resolution without span representations}.
\newblock In \emph{Proceedings of the 59th Annual Meeting of the Association
  for Computational Linguistics and the 11th International Joint Conference on
  Natural Language Processing (Volume 2: Short Papers)}, pages 14--19, Online.
  Association for Computational Linguistics.

\bibitem[{Krogh and Vedelsby(1994)}]{krogh1994neural}
Anders Krogh and Jesper Vedelsby. 1994.
\newblock Neural network ensembles, cross validation, and active learning.
\newblock \emph{Advances in neural information processing systems}, 7.

\bibitem[{Kudo and Richardson(2018)}]{kudo2018sentencepiece}
Taku Kudo and John Richardson. 2018.
\newblock Sentencepiece: A simple and language independent subword tokenizer
  and detokenizer for neural text processing.
\newblock \emph{arXiv preprint arXiv:1808.06226}.

\bibitem[{Kundu et~al.(2018)Kundu, Sil, Florian, and
  Hamza}]{kunduetal2018neural}
Gourab Kundu, Avi Sil, Radu Florian, and Wael Hamza. 2018.
\newblock \href {https://doi.org/10.18653/v1/P18-2063} {Neural cross-lingual
  coreference resolution and its application to entity linking}.
\newblock In \emph{Proceedings of the 56th Annual Meeting of the Association
  for Computational Linguistics (Volume 2: Short Papers)}, pages 395--400,
  Melbourne, Australia. Association for Computational Linguistics.

\bibitem[{Kurita et~al.(2019)Kurita, Vyas, Pareek, Black, and
  Tsvetkov}]{kuritaetal2019measuring}
Keita Kurita, Nidhi Vyas, Ayush Pareek, Alan~W Black, and Yulia Tsvetkov. 2019.
\newblock \href {https://doi.org/10.18653/v1/W19-3823} {Measuring bias in
  contextualized word representations}.
\newblock In \emph{Proceedings of the First Workshop on Gender Bias in Natural
  Language Processing}, pages 166--172, Florence, Italy. Association for
  Computational Linguistics.

\bibitem[{Lai et~al.(2021)Lai, Bui, and Kim}]{lai2021end}
Tuan~Manh Lai, Trung Bui, and Doo~Soon Kim. 2021.
\newblock End-to-end neural coreference resolution revisited: A simple yet
  effective baseline.
\newblock \emph{arXiv preprint arXiv:2107.01700}.

\bibitem[{Lai et~al.(2020)Lai, Tran, Bui, and Kihara}]{lai2020simple}
Tuan~Manh Lai, Quan~Hung Tran, Trung Bui, and Daisuke Kihara. 2020.
\newblock A simple but effective bert model for dialog state tracking on
  resource-limited systems.
\newblock In \emph{ICASSP 2020-2020 IEEE International Conference on Acoustics,
  Speech and Signal Processing (ICASSP)}, pages 8034--8038. IEEE.

\bibitem[{Lan et~al.(2020)Lan, Chen, Xu, and Ritter}]{lanetal2020empirical}
Wuwei Lan, Yang Chen, Wei Xu, and Alan Ritter. 2020.
\newblock \href {https://doi.org/10.18653/v1/2020.emnlp-main.382} {An empirical
  study of pre-trained transformers for {A}rabic information extraction}.
\newblock In \emph{Proceedings of the 2020 Conference on Empirical Methods in
  Natural Language Processing (EMNLP)}, pages 4727--4734, Online. Association
  for Computational Linguistics.

\bibitem[{Lee et~al.(2013)Lee, Chang, Peirsman, Chambers, Surdeanu, and
  Jurafsky}]{10.1162/COLI_a_00152}
Heeyoung Lee, Angel Chang, Yves Peirsman, Nathanael Chambers, Mihai Surdeanu,
  and Dan Jurafsky. 2013.
\newblock \href {https://doi.org/10.1162/COLI_a_00152} {{Deterministic
  Coreference Resolution Based on Entity-Centric, Precision-Ranked Rules}}.
\newblock \emph{Computational Linguistics}, 39(4):885--916.

\bibitem[{Lee et~al.(2017)Lee, He, Lewis, and Zettlemoyer}]{leeetal2017end}
Kenton Lee, Luheng He, Mike Lewis, and Luke Zettlemoyer. 2017.
\newblock \href {https://doi.org/10.18653/v1/D17-1018} {End-to-end neural
  coreference resolution}.
\newblock In \emph{Proceedings of the 2017 Conference on Empirical Methods in
  Natural Language Processing}, pages 188--197, Copenhagen, Denmark.
  Association for Computational Linguistics.

\bibitem[{Lee et~al.(2018)Lee, He, and Zettlemoyer}]{leeetal2018higher}
Kenton Lee, Luheng He, and Luke Zettlemoyer. 2018.
\newblock \href {https://doi.org/10.18653/v1/N18-2108} {Higher-order
  coreference resolution with coarse-to-fine inference}.
\newblock In \emph{Proceedings of the 2018 Conference of the North {A}merican
  Chapter of the Association for Computational Linguistics: Human Language
  Technologies, Volume 2 (Short Papers)}, pages 687--692, New Orleans,
  Louisiana. Association for Computational Linguistics.

\bibitem[{Li et~al.(2021)Li, Yavuz, Chen, and Yan}]{lietal2021taskadaptive}
Shiyang Li, Semih Yavuz, Wenhu Chen, and Xifeng Yan. 2021.
\newblock \href {https://doi.org/10.18653/v1/2021.findings-emnlp.86}
  {Task-adaptive pre-training and self-training are complementary for natural
  language understanding}.
\newblock In \emph{Findings of the Association for Computational Linguistics:
  EMNLP 2021}, pages 1006--1015, Punta Cana, Dominican Republic. Association
  for Computational Linguistics.

\bibitem[{Ling et~al.(2015)Ling, Singh, and Weld}]{lingetal2015design}
Xiao Ling, Sameer Singh, and Daniel~S. Weld. 2015.
\newblock \href {https://doi.org/10.1162/tacl_a_00141} {Design challenges for
  entity linking}.
\newblock \emph{Transactions of the Association for Computational Linguistics},
  3:315--328.

\bibitem[{Liu et~al.(2019)Liu, He, Chen, and Gao}]{Liu2019MultiTaskDN}
Xiaodong Liu, Pengcheng He, Weizhu Chen, and Jianfeng Gao. 2019.
\newblock Multi-task deep neural networks for natural language understanding.
\newblock In \emph{ACL}.

\bibitem[{Luo(2005)}]{Luo2005OnCR}
Xiaoqiang Luo. 2005.
\newblock On coreference resolution performance metrics.
\newblock In \emph{HLT}.

\bibitem[{Luo and Zitouni(2005)}]{luozitouni2005multi}
Xiaoqiang Luo and Imed Zitouni. 2005.
\newblock \href {https://aclanthology.org/H05-1083} {Multi-lingual coreference
  resolution with syntactic features}.
\newblock In \emph{Proceedings of Human Language Technology Conference and
  Conference on Empirical Methods in Natural Language Processing}, pages
  660--667, Vancouver, British Columbia, Canada. Association for Computational
  Linguistics.

\bibitem[{Melville and Mooney(2003)}]{Melville2003ConstructingDC}
Prem Melville and Raymond~J. Mooney. 2003.
\newblock Constructing diverse classifier ensembles using artificial training
  examples.
\newblock In \emph{IJCAI}.

\bibitem[{Min(2021)}]{min2021exploring}
Bonan Min. 2021.
\newblock \href {https://doi.org/10.18653/v1/2021.crac-1.10} {Exploring
  pre-trained transformers and bilingual transfer learning for {A}rabic
  coreference resolution}.
\newblock In \emph{Proceedings of the Fourth Workshop on Computational Models
  of Reference, Anaphora and Coreference}, pages 94--99, Punta Cana, Dominican
  Republic. Association for Computational Linguistics.

\bibitem[{Musgrave et~al.(2020)Musgrave, Belongie, and
  Lim}]{PyTorchMetricLearning}
Kevin Musgrave, Serge Belongie, and Ser-Nam Lim. 2020.
\newblock \href {http://arxiv.org/abs/2008.09164} {Pytorch metric learning}.

\bibitem[{Ng(2010)}]{ng2010supervised}
Vincent Ng. 2010.
\newblock \href {https://aclanthology.org/P10-1142} {Supervised noun phrase
  coreference research: The first fifteen years}.
\newblock In \emph{Proceedings of the 48th Annual Meeting of the Association
  for Computational Linguistics}, pages 1396--1411, Uppsala, Sweden.
  Association for Computational Linguistics.

\bibitem[{Ng(2017)}]{Ng17a}
Vincent Ng. 2017.
\newblock Machine learning for entity coreference resolution: A retrospective
  look at two decades of research.
\newblock In \emph{Proceedings of the 31st AAAI Conference on Artificial
  Intelligence}, pages 4877--4884.

\bibitem[{Nothman et~al.(2013)Nothman, Ringland, Radford, Murphy, and
  Curran}]{Nothman2013LearningMN}
Joel Nothman, Nicky Ringland, Will Radford, Tara Murphy, and James~R. Curran.
  2013.
\newblock Learning multilingual named entity recognition from wikipedia.
\newblock \emph{Artif. Intell.}, 194:151--175.

\bibitem[{Pan et~al.(2017)Pan, Zhang, May, Nothman, Knight, and
  Ji}]{panetal2017cross}
Xiaoman Pan, Boliang Zhang, Jonathan May, Joel Nothman, Kevin Knight, and Heng
  Ji. 2017.
\newblock \href {https://doi.org/10.18653/v1/P17-1178} {Cross-lingual name
  tagging and linking for 282 languages}.
\newblock In \emph{Proceedings of the 55th Annual Meeting of the Association
  for Computational Linguistics (Volume 1: Long Papers)}, pages 1946--1958,
  Vancouver, Canada. Association for Computational Linguistics.

\bibitem[{Paszke et~al.(2019)Paszke, Gross, Massa, Lerer, Bradbury, Chanan,
  Killeen, Lin, Gimelshein, Antiga, Desmaison, K{\"o}pf, Yang, DeVito, Raison,
  Tejani, Chilamkurthy, Steiner, Fang, Bai, and Chintala}]{Paszke2019PyTorchAI}
Adam Paszke, S.~Gross, Francisco Massa, A.~Lerer, J.~Bradbury, G.~Chanan,
  T.~Killeen, Z.~Lin, N.~Gimelshein, L.~Antiga, Alban Desmaison, Andreas
  K{\"o}pf, E.~Yang, Zach DeVito, Martin Raison, Alykhan Tejani, Sasank
  Chilamkurthy, B.~Steiner, Lu~Fang, Junjie Bai, and Soumith Chintala. 2019.
\newblock Pytorch: An imperative style, high-performance deep learning library.
\newblock In \emph{NeurIPS}.

\bibitem[{Peters et~al.(2018)Peters, Neumann, Iyyer, Gardner, Clark, Lee, and
  Zettlemoyer}]{petersetal2018deep}
Matthew~E. Peters, Mark Neumann, Mohit Iyyer, Matt Gardner, Christopher Clark,
  Kenton Lee, and Luke Zettlemoyer. 2018.
\newblock \href {https://doi.org/10.18653/v1/N18-1202} {Deep contextualized
  word representations}.
\newblock In \emph{Proceedings of the 2018 Conference of the North {A}merican
  Chapter of the Association for Computational Linguistics: Human Language
  Technologies, Volume 1 (Long Papers)}, pages 2227--2237, New Orleans,
  Louisiana. Association for Computational Linguistics.

\bibitem[{Poot and van Cranenburgh(2020)}]{pootvancranenburgh2020benchmark}
Corb{\`e}n Poot and Andreas van Cranenburgh. 2020.
\newblock \href {https://aclanthology.org/2020.crac-1.9} {A benchmark of
  rule-based and neural coreference resolution in {D}utch novels and news}.
\newblock In \emph{Proceedings of the Third Workshop on Computational Models of
  Reference, Anaphora and Coreference}, pages 79--90, Barcelona, Spain
  (online). Association for Computational Linguistics.

\bibitem[{Pradhan et~al.(2012)Pradhan, Moschitti, Xue, Uryupina, and
  Zhang}]{pradhanetal2012conll}
Sameer Pradhan, Alessandro Moschitti, Nianwen Xue, Olga Uryupina, and Yuchen
  Zhang. 2012.
\newblock \href {https://aclanthology.org/W12-4501} {{C}o{NLL}-2012 shared
  task: Modeling multilingual unrestricted coreference in {O}nto{N}otes}.
\newblock In \emph{Joint Conference on {EMNLP} and {C}o{NLL} - Shared Task},
  pages 1--40, Jeju Island, Korea. Association for Computational Linguistics.

\bibitem[{Pra{\v{z}}{\'a}k et~al.(2021)Pra{\v{z}}{\'a}k, Konop{\'\i}k, and
  Sido}]{prazaketal2021multilingual}
Ond{\v{r}}ej Pra{\v{z}}{\'a}k, Miloslav Konop{\'\i}k, and Jakub Sido. 2021.
\newblock \href {https://aclanthology.org/2021.ranlp-1.125} {Multilingual
  coreference resolution with harmonized annotations}.
\newblock In \emph{Proceedings of the International Conference on Recent
  Advances in Natural Language Processing (RANLP 2021)}, pages 1119--1123, Held
  Online. INCOMA Ltd.

\bibitem[{Raghunathan et~al.(2010)Raghunathan, Lee, Rangarajan, Chambers,
  Surdeanu, Jurafsky, and Manning}]{raghunathanetal2010multi}
Karthik Raghunathan, Heeyoung Lee, Sudarshan Rangarajan, Nathanael Chambers,
  Mihai Surdeanu, Dan Jurafsky, and Christopher Manning. 2010.
\newblock \href {https://aclanthology.org/D10-1048} {A multi-pass sieve for
  coreference resolution}.
\newblock In \emph{Proceedings of the 2010 Conference on Empirical Methods in
  Natural Language Processing}, pages 492--501, Cambridge, MA. Association for
  Computational Linguistics.

\bibitem[{Recasens et~al.(2010)Recasens, M{\`a}rquez, Sapena, Mart{\'\i},
  Taul{\'e}, Hoste, Poesio, and Versley}]{recasensetal2010semeval}
Marta Recasens, Llu{\'\i}s M{\`a}rquez, Emili Sapena, M.~Ant{\`o}nia
  Mart{\'\i}, Mariona Taul{\'e}, V{\'e}ronique Hoste, Massimo Poesio, and
  Yannick Versley. 2010.
\newblock \href {https://aclanthology.org/S10-1001} {{S}em{E}val-2010 task 1:
  Coreference resolution in multiple languages}.
\newblock In \emph{Proceedings of the 5th International Workshop on Semantic
  Evaluation}, pages 1--8, Uppsala, Sweden. Association for Computational
  Linguistics.

\bibitem[{Sheng et~al.(2021)Sheng, Chang, Natarajan, and
  Peng}]{sheng2021societal}
Emily Sheng, Kai-Wei Chang, Prem Natarajan, and Nanyun Peng. 2021.
\newblock Societal biases in language generation: Progress and challenges.
\newblock In \emph{ACL}.

\bibitem[{Singh et~al.(2012)Singh, Subramanya, Pereira, and
  McCallum}]{Singh2012WikilinksAL}
Sameer Singh, Amarnag Subramanya, Fernando Pereira, and Andrew McCallum. 2012.
\newblock Wikilinks: A large-scale cross-document coreference corpus labeled
  via links to wikipedia.
\newblock \emph{University of Massachusetts, Amherst, Tech. Rep. UM-CS-2012},
  15.

\bibitem[{van Cranenburgh(2019)}]{vanCranenburgh2019ADC}
Andreas van Cranenburgh. 2019.
\newblock A dutch coreference resolution system with an evaluation on literary
  fiction.
\newblock \emph{Computational Linguistics in the Netherlands Journal},
  9:27--54.

\bibitem[{Vilain et~al.(1995)Vilain, Burger, Aberdeen, Connolly, and
  Hirschman}]{Vilain1995AMC}
Marc~B. Vilain, John~D. Burger, John~S. Aberdeen, Dennis Connolly, and Lynette
  Hirschman. 1995.
\newblock A model-theoretic coreference scoring scheme.
\newblock In \emph{MUC}.

\bibitem[{Webster et~al.(2018)Webster, Recasens, Axelrod, and
  Baldridge}]{gapdataset}
Kellie Webster, Marta Recasens, Vera Axelrod, and Jason Baldridge. 2018.
\newblock \href {https://doi.org/10.1162/tacl_a_00240} {{Mind the GAP: A
  Balanced Corpus of Gendered Ambiguous Pronouns}}.
\newblock \emph{Transactions of the Association for Computational Linguistics},
  6:605--617.

\bibitem[{Wolf et~al.(2020)Wolf, Debut, Sanh, Chaumond, Delangue, Moi, Cistac,
  Rault, Louf, Funtowicz, Davison, Shleifer, von Platen, Ma, Jernite, Plu, Xu,
  Le~Scao, Gugger, Drame, Lhoest, and Rush}]{wolfetal2020transformers}
Thomas Wolf, Lysandre Debut, Victor Sanh, Julien Chaumond, Clement Delangue,
  Anthony Moi, Pierric Cistac, Tim Rault, Remi Louf, Morgan Funtowicz, Joe
  Davison, Sam Shleifer, Patrick von Platen, Clara Ma, Yacine Jernite, Julien
  Plu, Canwen Xu, Teven Le~Scao, Sylvain Gugger, Mariama Drame, Quentin Lhoest,
  and Alexander Rush. 2020.
\newblock \href {https://www.aclweb.org/anthology/2020.emnlp-demos.6}
  {Transformers: State-of-the-art natural language processing}.
\newblock In \emph{Proceedings of the 2020 Conference on Empirical Methods in
  Natural Language Processing: System Demonstrations}, pages 38--45, Online.
  Association for Computational Linguistics.

\bibitem[{Wu et~al.(2020{\natexlab{a}})Wu, Petroni, Josifoski, Riedel, and
  Zettlemoyer}]{wu2020scalable}
Ledell Wu, Fabio Petroni, Martin Josifoski, Sebastian Riedel, and Luke
  Zettlemoyer. 2020{\natexlab{a}}.
\newblock \href {https://doi.org/10.18653/v1/2020.emnlp-main.519} {Scalable
  zero-shot entity linking with dense entity retrieval}.
\newblock In \emph{Proceedings of the 2020 Conference on Empirical Methods in
  Natural Language Processing (EMNLP)}, pages 6397--6407, Online. Association
  for Computational Linguistics.

\bibitem[{Wu et~al.(2020{\natexlab{b}})Wu, Wang, Yuan, Wu, and
  Li}]{wuetal2020corefqa}
Wei Wu, Fei Wang, Arianna Yuan, Fei Wu, and Jiwei Li. 2020{\natexlab{b}}.
\newblock \href {https://doi.org/10.18653/v1/2020.acl-main.622} {{C}oref{QA}:
  Coreference resolution as query-based span prediction}.
\newblock In \emph{Proceedings of the 58th Annual Meeting of the Association
  for Computational Linguistics}, pages 6953--6963, Online. Association for
  Computational Linguistics.

\bibitem[{Xia et~al.(2020)Xia, Sedoc, and Van~Durme}]{xiaetal2020incremental}
Patrick Xia, Jo{\~a}o Sedoc, and Benjamin Van~Durme. 2020.
\newblock \href {https://doi.org/10.18653/v1/2020.emnlp-main.695} {Incremental
  neural coreference resolution in constant memory}.
\newblock In \emph{Proceedings of the 2020 Conference on Empirical Methods in
  Natural Language Processing (EMNLP)}, pages 8617--8624, Online. Association
  for Computational Linguistics.

\bibitem[{Xia and Van~Durme(2021)}]{xiavandurme2021moving}
Patrick Xia and Benjamin Van~Durme. 2021.
\newblock \href {https://doi.org/10.18653/v1/2021.emnlp-main.425} {Moving on
  from {O}nto{N}otes: Coreference resolution model transfer}.
\newblock In \emph{Proceedings of the 2021 Conference on Empirical Methods in
  Natural Language Processing}, pages 5241--5256, Online and Punta Cana,
  Dominican Republic. Association for Computational Linguistics.

\bibitem[{Xu and Choi(2020)}]{xuchoi2020revealing}
Liyan Xu and Jinho~D. Choi. 2020.
\newblock \href {https://doi.org/10.18653/v1/2020.emnlp-main.686} {Revealing
  the myth of higher-order inference in coreference resolution}.
\newblock In \emph{Proceedings of the 2020 Conference on Empirical Methods in
  Natural Language Processing (EMNLP)}, pages 8527--8533, Online. Association
  for Computational Linguistics.

\bibitem[{Yang et~al.(2012)Yang, Mao, Xiang, Tsang, Chai, and
  Chieu}]{yangetal2012domain}
Jian~Bo Yang, Qi~Mao, Qiao~Liang Xiang, Ivor Wai-Hung Tsang, Kian Ming~Adam
  Chai, and Hai~Leong Chieu. 2012.
\newblock \href {https://aclanthology.org/D12-1068} {Domain adaptation for
  coreference resolution: An adaptive ensemble approach}.
\newblock In \emph{Proceedings of the 2012 Joint Conference on Empirical
  Methods in Natural Language Processing and Computational Natural Language
  Learning}, pages 744--753, Jeju Island, Korea. Association for Computational
  Linguistics.

\bibitem[{Zhao et~al.(2018)Zhao, Wang, Yatskar, Ordonez, and
  Chang}]{zhaoetal2018gender}
Jieyu Zhao, Tianlu Wang, Mark Yatskar, Vicente Ordonez, and Kai-Wei Chang.
  2018.
\newblock \href {https://doi.org/10.18653/v1/N18-2003} {Gender bias in
  coreference resolution: Evaluation and debiasing methods}.
\newblock In \emph{Proceedings of the 2018 Conference of the North {A}merican
  Chapter of the Association for Computational Linguistics: Human Language
  Technologies, Volume 2 (Short Papers)}, pages 15--20, New Orleans, Louisiana.
  Association for Computational Linguistics.

\end{thebibliography}
\bibliographystyle{acl_natbib}

\appendix
\label{sec:appendix}
\section{Reproducibility Information}

In this section, we present the reproducibility information of our paper.

\paragraph{Implementation Dependencies Libraries} Pytorch 1.11.0 \cite{Paszke2019PyTorchAI}, Transformers 4.17.0 \cite{wolfetal2020transformers}, SentencePiece 0.1.96 \cite{kudo2018sentencepiece}, PyTorch Metric Learning \cite{PyTorchMetricLearning}.

\paragraph{Computing Infrastructure} The experiments were conducted on a server with Intel(R) Xeon(R) Gold 5120 CPU @ 2.20GHz and NVIDIA Tesla V100 GPUs. GPU memory is 16G.

\paragraph{Number of Model Parameters} When the target dataset is OntoNotes Arabic, we use GigaBERT \cite{lanetal2020empirical} as the Transformer encoder. GigaBERT has about 125M parameters.

When the target dataset is SemEval Dutch or SemEval Spanish, we use the base version of XLM-R (i.e., \texttt{xlm-roberta-base}) \cite{Conneau2020UnsupervisedCR}. \texttt{xlm-roberta-base} has about 278M parameters.

\paragraph{Hyperparameters} The information about the hyperparameters is available in the main paper.

\paragraph{Expected Validation Performance} We report the validation performance of the ensemble \textit{[Continued Training $\oplus$ Joint Training $\oplus$ Wikipedia Pre-Training]}.

The validation F1 score of the ensemble for Arabic coreference resolution is 66.60\%. The total time needed for the evaluation is about 1 minute and 19 seconds.

The validation F1 score of the ensemble for Dutch coreference resolution is 57.81\%. The total time needed for the evaluation is about 20 seconds.

The validation F1 score of the ensemble for Spanish coreference resolution is 75.73\%. The total time needed for the evaluation is about 1 minute and 42 seconds.

\end{document}